\PassOptionsToPackage{unicode}{hyperref}
\PassOptionsToPackage{hyphens}{url}
\documentclass[
]{article}
\usepackage{amsmath,amssymb}
\usepackage{iftex}
\ifPDFTeX
  \usepackage[T1]{fontenc}
  \usepackage[utf8]{inputenc}
  \usepackage{textcomp} 
\else 
  \usepackage{unicode-math} 
  \defaultfontfeatures{Scale=MatchLowercase}
  \defaultfontfeatures[\rmfamily]{Ligatures=TeX,Scale=1}
\fi
\usepackage{lmodern}
\ifPDFTeX\else
\fi
\IfFileExists{upquote.sty}{\usepackage{upquote}}{}
\IfFileExists{microtype.sty}{
  \usepackage[]{microtype}
  \UseMicrotypeSet[protrusion]{basicmath} 
}{}
\makeatletter
\@ifundefined{KOMAClassName}{
  \IfFileExists{parskip.sty}{%
    \usepackage{parskip}
  }{
    \setlength{\parindent}{0pt}
    \setlength{\parskip}{6pt plus 2pt minus 1pt}}
}{
  \KOMAoptions{parskip=half}}
\makeatother
\usepackage{xcolor}
\usepackage[margin=1in]{geometry}
\usepackage{color}
\usepackage{fancyvrb}

\DefineVerbatimEnvironment{Highlighting}{Verbatim}{commandchars=\\\{\}}
\usepackage{framed}
\definecolor{shadecolor}{RGB}{248,248,248}
\newenvironment{Shaded}{\begin{snugshade}}{\end{snugshade}}

\newcommand{\BuiltInTok}[1]{#1}

\newcommand{\CommentTok}[1]{\textcolor[rgb]{0.56,0.35,0.01}{\textit{#1}}}

\newcommand{\ControlFlowTok}[1]{\textcolor[rgb]{0.13,0.29,0.53}{\textbf{#1}}}

\newcommand{\DecValTok}[1]{\textcolor[rgb]{0.00,0.00,0.81}{#1}}

\newcommand{\FloatTok}[1]{\textcolor[rgb]{0.00,0.00,0.81}{#1}}

\newcommand{\KeywordTok}[1]{\textcolor[rgb]{0.13,0.29,0.53}{\textbf{#1}}}
\newcommand{\NormalTok}[1]{#1}
\newcommand{\OperatorTok}[1]{\textcolor[rgb]{0.81,0.36,0.00}{\textbf{#1}}}

\newcommand{\StringTok}[1]{\textcolor[rgb]{0.31,0.60,0.02}{#1}}

\usepackage{longtable,booktabs,array}
\usepackage{calc} 
\usepackage{etoolbox}
\makeatletter
\patchcmd\longtable{\par}{\if@noskipsec\mbox{}\fi\par}{}{}
\makeatother
\IfFileExists{footnotehyper.sty}{\usepackage{footnotehyper}}{\usepackage{footnote}}
\makesavenoteenv{longtable}
\usepackage{graphicx}
\makeatletter
\def\maxwidth{\ifdim\Gin@nat@width>\linewidth\linewidth\else\Gin@nat@width\fi}
\def\maxheight{\ifdim\Gin@nat@height>\textheight\textheight\else\Gin@nat@height\fi}
\makeatother
\setkeys{Gin}{width=\maxwidth,height=\maxheight,keepaspectratio}
\makeatletter
\def\fps@figure{htbp}
\makeatother
\providecommand{\tightlist}{%
  \setlength{\itemsep}{0pt}\setlength{\parskip}{0pt}}
\ifLuaTeX
  \usepackage{selnolig}  
\fi
\usepackage{bookmark}
\IfFileExists{xurl.sty}{\usepackage{xurl}}{} 
\hypersetup{
  pdftitle={AI-ANNE: (A) (N)eural (N)et for (E)xploration},
  pdfauthor={a Working Paper and Manual by Dennis Klinkhammer},
  hidelinks,
  pdfcreator={LaTeX via pandoc}}

\title{AI-ANNE: (A) (N)eural (N)et for (E)xploration}
\usepackage{etoolbox}
\makeatletter
\providecommand{\subtitle}[1]{
  \apptocmd{\@title}{\par {\large #1 \par}}{}{}
}
\makeatother
\subtitle{Transferring Deep Learning Models onto Microcontrollers and
Embedded Systems}
\author{a Working Paper and Manual by Dennis Klinkhammer}
\date{}

\begin{document}
\maketitle

\subsection{Abstract (ENG)}\label{abstract-eng}

This working paper explores the integration of neural networks onto
resource-constrained embedded systems like a Raspberry Pi Pico /
Raspberry Pi Pico 2. A TinyML aproach transfers neural networks directly
on these microcontrollers, enabling real-time, low-latency, and
energy-efficient inference while maintaining data privacy. Therefore,
AI-ANNE: (A) (N)eural (N)et for (E)xploration will be presented, which
facilitates the transfer of pre-trained models from high-performance
platforms like TensorFlow and Keras onto microcontrollers, using a
lightweight programming language like MicroPython. This approach
demonstrates how neural network architectures, such as neurons, layers,
density and activation functions can be implemented in MicroPython in
order to deal with the computational limitations of embedded systems.
Based on the Raspberry Pi Pico / Raspberry Pi Pico 2, two different
neural networks on microcontrollers are presented for an example of data
classification. As an further application example, such a
microcontroller can be used for condition monitoring, where immediate
corrective measures are triggered on the basis of sensor data. Overall,
this working paper presents a very easy-to-implement way of using neural
networks on energy-efficient devices such as microcontrollers. This
makes AI-ANNE: (A) (N)eural (N)et for (E)xploration not only suited for
practical use, but also as an educational tool with clear insights into
how neural networks operate.

\subsection{Abstract (GER)}\label{abstract-ger}

Dieser vorläufige Artikel befasst sich mit der Integration neuronaler
Netze in ressourcenlimitierte und eingebettete Systeme wie den Raspberry
Pi Pico / Raspberry Pi Pico 2. Ein TinyML-Ansatz überträgt dabei
neuronale Netze direkt auf einen Mikrocontroller und ermöglicht so eine
latenzarme und energieeffiziente Datenanalyse bei gleichzeitiger
Sicherung sensibler Daten. Zu diesem Zweck wird KI-ENNA: (E)in
(N)euronales (N)etz zum (A)usprobieren vorgestellt, das die Übertragung
von vortrainierten und rechenintensiven Modellen auf Basis von
TensorFlow und Keras auf Mikrocontroller unter Verwendung einer
leichtgewichtigen Programmiersprache wie MicroPython ermöglicht. Dieser
Ansatz verdeutlicht dabei, wie die den neuronalen Netzen
zugrundeliegende Architektur in Form von Neuronen, Schichten, Dichte und
Aktivierungsfunktionen in MicroPython implementiert werden kann, um
unter den Ressourcenlimitationen von eingebetteten Systemen
funktionsfähig zu sein. Auf Grundlage des Raspberry Pi Pico / Raspberry
Pi Pico 2 werden zwei verschiedene neuronale Netze auf Mikrocontrollern
für ein Beispiel zur Datenklassifizierung vorgestellt. In der Praxis
wäre mit einem solchen Microcontroller darüber hinaus eine
Zustandsüberwachung möglich, bei denen auf Basis von Sensordaten
sofortige Korrekturmaßnahmen ausgelöst werden. Dadurch stellt dieser
vorläufige Beitrag insgesamt eine sehr leicht umzusetzende Möglichkeit
vor, wie neuronale Netze auf energieeffizienten Geräten wie
Mikrocontroller zur Anwendung gebracht werden können. Dadurch ist
KI-ENNA: (E)in (N)euronales (N)etz zum (A)usprobieren nicht nur eine
Option für die Praxis, sondern gleichermaßen ein didaktisches Tool mit
anschaulichen Einblicken in die Funkktionsweise neuronaler Netze.

\subsubsection{Keywords}\label{keywords}

TinyML, EdgeAI, Microcontroller, Embedded Systems, Machine Learning,
Deep Learning

\section{(I) Introduction}\label{i-introduction}

\subsection{Artificial Intelligence for
Microcontrollers}\label{artificial-intelligence-for-microcontrollers}

Machine Learning and Deep Learning are increasingly driving innovation
across various fields (LeCun et al.~2015), with a notable expansion into
embedded systems, bringing their capabilities closer to data sources.
This trend, known as TinyML, is especially prominent in microcontrollers
and Internet of Things devices. TinyML offers several advantages over
cloud-based artificial intelligence, including improved data privacy,
lower processing latency, energy efficiency, and reduced dependency on
connectivity (Ray 2022). One key application of TinyML is in condition
monitoring, where neural networks can be used to detect anomalies
directly within sensors so that immediate corrective actions can be
taken automatically (Cioffi et al.~2020). Therefore, when these
microcontrollers are connected to microscopes in medical diagnostics or
machines for industrial production, they are referred to as embedded
systems.

While many high-performance frameworks like TensorFlow and Keras for
Python are designed for powerful hardware such as GPUs, these frameworks
are unsuitable for embedded systems due to their limited computational
resources (Delnevo et al.~2023). To address this issue, the architecture
of neural networks need to be reconstructed in a leightweight
programming language like MicroPython. Thus, neural networks trained on
high-performance hardware can be transferred onto resource-constrained
devices like microcontrollers while achieving the same outputs and
results (Ray 2022). However, training or developing neural networks
directly on embedded systems remains a complex challenge. AIfES:
(A)rtificial (I)ntelligence (f)or (E)mbedded (S)ystems, as presented
below as a comparable approach, has shown initial success here (Wulfert
et al.~2024).

This working paper presents AI-ANNE: (A) (N)eural (N)et for
(E)xploration, a new method for transferring pre-trained neural networks
onto a microcontroller. For this purpose, neurons, layers, density and
activation functions as the underlying foundation of neural networks
were coded in MicroPython in order to be able to reconstruct the
original neural networks trained with TensorFlow and Keras. How the
transfer works and a suitable application example are part of this
working paper. The underlying foundation of neural networks and their
counterparts in MicroPython are presented successively.

\subsection{Comparable Approaches and Added
Value}\label{comparable-approaches-and-added-value}

AIfES: (A)rtificial (I)ntelligence (f)or (E)mbedded (S)ystems is a
flexible software framework designed by Fraunhofer to run deep learning
models on small, low-power devices like microcontrollers (Wulfert et
al.~2024). It simplifies the process of building, training, and running
models directly on these devices, without needing powerful external
systems. Users can customize the framework to fit their needs by
choosing different model components, like types of layers or how data is
processed. AIfES: (A)rtificial (I)ntelligence (f)or (E)mbedded (S)ystems
can run pre-trained models and train new ones on the device itself,
saving energy and protecting privacy (Wulfert et al.~2024). It
outperforms similar tools in terms of speed and memory usage for certain
tasks. Future improvements will focus on making it even more efficient
and supporting new types of deep learning models.

AI-ANNE: (A) (N)eural (N)et for (E)xploration is a similar approach and,
as an open framework, enables the flexible expansion of the underlying
activation functions in order to explore their performance while
simultaneously the number of neurons and layers can be adjusted easily
in MicroPython. This flexibility can also be used for fine-tuning
directly on the microcontroller. As a result, AI-ANNE: (A) (N)eural
(N)et for (E)xploration allows the learning behavior of the neural
networks to be observed and creates a didactic value for its users. Two
neural networks are already pre-installed: One with six neurons in a
total of three layers and one with eight neurons in a total of four
layers. In the given example, the learning behavior of the various
neural networks can be investigated with the pre-installed IRIS dataset.
The transparent insight into the MicroPython code also opens up didactic
application potential. The interaction with the microcontroller takes
place via Thonny. All the necessary components are presented below.
\newpage

\section{(II) Requirements}\label{ii-requirements}

\subsection{Raspberry Pi Pico / Raspberry Pi Pico
2}\label{raspberry-pi-pico-raspberry-pi-pico-2}

The Raspberry Pi Pico is powered by the RP2040 microcontroller, which
features a dual-core ARM Cortex-M0+ processor running at 133 MHz, with
264 KB on-chip SRAM and 2 MB onboard flash memory. It offers a wide
range of connectivity options, including USB for power and data
transfer, up to two I2C, SPI, and UART interfaces for communication, as
well as 16 PWM channels for precise control of external devices.

The board also includes three 12-bit ADC channels for analog input, and
it supports a range of peripherals such as a real-time clock (RTC),
timers, and interrupt handling through a nested vectored interrupt
controller (NVIC). For power, the Raspberry Pi Pico operates on a
voltage range of 1.8V to 3.6V, with typical consumption between 20-100
mA, and can be powered via the micro-USB port or the VSYS pin for
external power sources like batteries or regulated supplies.

In 2024 an updated Raspberry Pi Pico 2 was introduced, which is powered
by the RP2350 microcontroller, which features a dual-core ARM Cortex-M33
processor running at 150MHz and 520 KB on-chip SRAM and 4 MB onboard
flash memory. Both can be operated with MicroPython. It can be assumed
that the Raspberry Pi Pico 2 can calculate larger datasets and more
complex neural networks with AI-ANNE: (A) (N)eural (N)et for
(E)xploration. Therefore, the use of a Raspberry Pi Pico 2 is
recommended for practical use; The use of the Raspberry Pi Pico might be
sufficient for educational purpose.

\subsection{MicroPython}\label{micropython}

MicroPython is an efficient, lightweight implementation of the Python
programming language designed to run on microcontrollers and embedded
systems with constrained resources (Delnevo et al.~2023). Unlike the
full Python environment, MicroPython is optimized to operate within the
memory and processing limits typical of small-scale devices, offering a
streamlined interpreter and a subset of Python's standard libraries.
MicroPython retains much of Python's high-level syntax and ease of use,
making it accessible to developers familiar with Python. It is
particularly well-suited for rapid prototyping, development, and
deployment of machine learning and neural network models on embedded
platforms, where resources such as memory, computational power, and
storage are limited.

In the context of neural network applications, MicroPython is often used
in edge computing scenarios, where deep learning models need to be
deployed directly onto microcontroller-based systems for real-time,
localized inference. Although MicroPython does not natively support the
extensive numerical libraries found in full Python (e.g., TensorFlow and
Keras), it is possible to reproduce the basic architecture of neural
networks with AI-ANNE: (A) (N)eural (N)et for (E)xploration in
MicroPython from scratch.

\subsection{Thonny}\label{thonny}

Thonny is a simple, user-friendly program that works on all major
computer systems. It makes it easy to connect with and program the
Raspberry Pi Pico / Raspberry Pi Pico 2. Thonny enables users to quickly
write and test code, manage files and fix any mistakes with helpful
tools. It can be described as a tool for those just starting with
MicroPython on microcontrollers and embedded systems. With Thonny,
AI-ANNE: (A) (N)eural (N)et for (E)xploration can simply be flashed onto
the microcontroller.

Thonny can be downloaded here: \url{https://thonny.org/} \newpage

\section{(III) Basic Knowledge}\label{iii-basic-knowledge}

\subsection{Architecture of Neural
Networks}\label{architecture-of-neural-networks}

A neural network is a computational model inspired by the way biological
neural systems process information. It consists of interconnected layers
of nodes, or neurons, which transform input data into output predictions
through a series of mathematical operations (LeCun et al.~2015). In the
context of implementing neural networks in MicroPython, the architecture
must be designed to operate within the resource constraints of embedded
systems (Ray 2022). Despite these constraints, the basic components of a
neural network - neurons, layers, density, and activation functions -
can still be effectively modeled (Sakr et al.~2021).

\subsubsection{a) Neurons}\label{a-neurons}

A neuron in a neural network is a computational unit that receives
inputs, applies a weight to each input, sums the weighted inputs, and
passes the result through an activation function to produce an output.
In MicroPython (Appendix - C), each neuron can be represented as a
mathematical operation involving inputs and weights, with the output
being calculated via simple matrix operations.

The general form of the neuron's computation can be expressed with \(y\)
as the output of the neuron, \(f\) as the activation function, \(x_i\)
as the inputs, \(w_i\) as the corresponding weights and \(b\) as the
bias:

\[ y = f(\sum_{i} w_i x_i + b) \]

In a MicroPython implementation, these calculations can be done using
basic array operations, making it computationally efficient for
small-scale neural networks on microcontrollers.

\subsubsection{b) Layers}\label{b-layers}

A neural network is typically structured as a series of layers, where
each layer consists of multiple neurons. There are three main types of
layers in a typical neural network architecture:

\begin{itemize}
\item
  Input Layer: This is the first layer of the network and receives the
  raw input data. Each neuron in the input layer corresponds to a
  feature in the input data. With AI-ANNE: (A) (N)eural (N)et for
  (E)xploration, this layer would typically read sensor values or
  external data directly.
\item
  Hidden Layers: Between the input and output layers, the network may
  contain one or more hidden layers. These layers are responsible for
  learning complex representations of the input data. Each hidden layer
  is composed of neurons that perform weighted sums of the outputs of
  the previous layer, followed by an activation function.
\item
  Output Layer: This is the final layer of the network, which produces
  the prediction or classification result. The output layer's structure
  depends on the problem being solved (e.g., binary classification,
  multi-class classification, logistic regression).
\end{itemize}

MicroPython can represent the layers as a list of lists (or arrays) of
neurons, where each element stores the weights, biases, and outputs of
the neurons in the respective layer.

\subsubsection{c) Density}\label{c-density}

The density of a layer refers to the number of neurons in that layer. In
a fully connected (dense) layer, each neuron in the current layer is
connected to all neurons in the previous layer. This is the most common
configuration in neural networks. For example: In a dense layer with 5
neurons, each neuron in the layer receives input from all neurons in the
previous layer, and the layer will contain 5 sets of weights and biases
to be learned during training. MicroPython can implement a dense layer
efficiently using matrix multiplication (Appendix - A, all of a sudden,
school math seems very practical!). Each layer's input and weight
matrices are multiplied together, followed by the addition of a bias
term, and the result is passed through an activation function.

\subsubsection{d) Activation Functions}\label{d-activation-functions}

An activation function determines whether a neuron should activate, so
that the neuron's output will be passed forward to the next layer. The
activation function introduces non-linearity into the network, allowing
it to learn complex patterns in the data. Common activation functions
include:

\begin{itemize}
\tightlist
\item
  Sigmoid: The Sigmoid function outputs values between 0 and 1 and is
  often used in binary classification tasks.
\end{itemize}

\[Sigmoid(x) = \frac{1}{1 + e^{-x}}\]

\begin{itemize}
\tightlist
\item
  ReLU (Rectified Linear Unit): ReLU outputs the input directly if it is
  positive; otherwise, it outputs zero. It is widely used in hidden
  layers due to its simplicity and efficiency in preventing vanishing
  gradients.
\end{itemize}

\[{ReLU}(x) = 
\begin{cases} 
x & \text{if } x \geq 0 \\
0 & \text{if } x < 0 
\end{cases}\]

\begin{itemize}
\tightlist
\item
  Leaky ReLU: The Leaky ReLU activation function is a variant of the
  traditional ReLU activation function, which addresses a common issue
  known as the ``dying ReLU'' problem. This problem occurs when neurons
  become inactive and stop learning because their output is always zero
  (when the input is negative). Leaky ReLU overcomes this by allowing a
  small, non-zero gradient when the input is negative. The key
  difference between ReLU and Leaky ReLU is that when \(x < 0\), instead
  of the output being zero (as in ReLU), the output is a small negative
  value. The parameter \(alpha\) controls the slope of this negative
  region, typically with small values such as 0.01 or 0.1.
\end{itemize}

\[{Leaky ReLU}(x) = 
\begin{cases} 
x & \text{if } x \geq 0 \\
\alpha x & \text{if } x < 0 
\end{cases}\]

\begin{itemize}
\tightlist
\item
  Tanh: The hyperbolic tangent function Tanh outputs values between -1
  and 1. It is similar to the sigmoid but has a wider output range,
  making it useful for some applications where the network needs to
  model both positive and negative values.
\end{itemize}

\[Tanh(x) = \frac{e^x - e^{-x}}{e^x + e^{-x}}\]

\begin{itemize}
\tightlist
\item
  Softmax: Often used in the output layer of classification networks,
  the Softmax function normalizes the output to produce a probability
  distribution across multiple classes.
\end{itemize}

\[{Softmax}(x_i) = \frac{e^{x_i}}{\sum_{j} e^{x_j}}\]

In MicroPython, these activation functions can be implemented as simple
functions, leveraging MicroPython's built-in math library (Appendix -
B). Given the computational constraints of embedded devices, it is
crucial to choose lightweight activation functions like ReLU, which
avoid the more computationally expensive exponentiation operations used
in Sigmoid or Tanh. AI-ANNE: (A) (N)eural (N)et for (E)xploration passes
data through the neural network layer by layer, and each neuron's output
is computed based on the inputs and weights. In a MicroPython
environment, due to the limited processing power and memory, a smaller
network with fewer layers and neurons might be preferred, and training
could be done in advance on a more powerful system before deployment to
the embedded device. \newpage

\subsubsection{e) Weights and Biases}\label{e-weights-and-biases}

In neural networks, weights and biases are key parameters that the model
learns during the training process to make accurate predictions.

Weights are the parameters that scale the input values as they pass
through the network. They determine the strength of the connections
between neurons in adjacent layers. Each connection between two neurons
has its own weight, and the value of this weight influences how much the
input from the previous layer affects the output of the current layer.
In mathematical terms, if a neuron receives an input \(x\) and its
corresponding weight is \(w\), the contribution of that input to the
neuron's output is \(w*x\). Weights are adjusted during training using
optimization techniques, such as gradient descent, to minimize the
model's prediction error.

Biases, on the other hand, allow the model to shift the output
independently of the weighted sum of inputs. They are added to the
weighted sum before it is passed through the activation function. This
allows the network to better model complex relationships in the data by
shifting the activation function's threshold. For example, if the
weighted sum of inputs to a neuron is represented as \(z=w*x+b\), the
bias term \(b\) shifts the output, enabling the model to learn more
effectively. Like the weights, biases are also learned during training.

\subsection{Confusion Matrix}\label{confusion-matrix}

A confusion matrix is like a scoreboard that indicates how well a neural
network is doing at making predictions. It's especially useful for
classification problems, where the goal is to assign items to
categories. For the purpose of binary classification a confusion matrix
can be a table with two rows and two columns. Each cell tells something
about the model's predictions compared to the actual results, which can
be interpreted as follows:

\begin{itemize}
\item
  True Positives (TP): The model correctly predicted the positive class.
\item
  True Negatives (TN): The model correctly predicted the negative class.
\item
  False Positives (FP): The model incorrectly predicted the positive
  class (Type I Error).
\item
  False Negatives (FN): The model incorrectly predicted the negative
  class (Type II Error).
\end{itemize}

The Accuracy of the neural network based prediction can be calculated as
a metric based on this. Accuracy as a percentage is a metric that
provides information about how often the model is right overall. Below
is an illustrative example (Tab. 1) of a confusion matrix with actual
values in the rows and predicted values in the columns:

\subsubsection{(Tab. 1) Confusion Matrix
Example}\label{tab.-1-confusion-matrix-example}

\begin{longtable}[]{@{}lll@{}}
\toprule\noalign{}
& Predicted Positive & Predicted Negative \\
\midrule\noalign{}
\endhead
\bottomrule\noalign{}
\endlastfoot
Actual Positive & 09.00 & 01.00 \\
Actual Negative & 00.00 & 10.00 \\
\end{longtable}

In this binary classification, \(09.00\) of the positive class are
classified correctly and \(01.00\) incorrectly (Type II Error).
Accordingly, \(10.00\) of the negative class are classified correctly
and \(00.00\) incorrectly. This results in an Accuracy of 95\% (19.00 /
20.00 = 00.95) and is based on the second application example of a
neural network with 6 neurons, which is presented below. In addition to
this metric, there are other metrics for the evaluation of neural
networks, e.g.~Precision, Recall and F1 Score. For this example and as
basic introduction to AI-ANNE: (A) (N)eural (N)et for (E)xploration, the
focus on accuracy shall suffice. \newpage

\section{(IV) Application Example}\label{iv-application-example}

\subsection{Binary Classsification}\label{binary-classsification}

The IRIS dataset is a widely recognized dataset in statistics, machine
learning and deep learning and often used for classification problems.
Introduced by the British biologist and statistician Ronald A. Fisher in
1936, it contains 150 instances of iris flowers, each with four
attributes that describe their physical characteristics. These
attributes include the sepal length, sepal width, petal length, and
petal width, all measured in centimeters.

Based on these four attributes, the objective of the dataset is to
classify each flower into one of three species: Setosa, Versicolor, and
Virginica. The IRIS dataset serves as an ideal example for demonstrating
and testing machine learning algorithms and neural networks as deep
learning models, particularly for classification tasks. Its manageable
size, clear distinctions between classes and straightforward nature make
it a popular choice for exploring classification techniques. However,
classifying some of the species, particularly Versicolor and Virginica,
can be challenging due to their overlapping characteristics, making the
task more complex for certain machine learning algorithms and neural
networks as deep learning models.

In the pre-installed example, AI-ANNE: (A) (N)eural (N)et for
(E)xploration was entrusted with the classification of Versicolor and
Virginica accordingly. The result will be a percentage, as presented
before as Accuracy.

\subsection{Solution with 8 Neurons}\label{solution-with-8-neurons}

First, with TensorFlow and Keras pre-trained weights and biases must be
transferred onto the microcontroller. The weights and biases of the
neural network with 8 neurons need to be coded in a specific order in
MicroPython (Tab. 2). The first layer, the so called input layer,
contains two neurons that process the input from the four independent
variables of the IRIS dataset. The weights \(w1\) are coded accordingly
with two columns and four rows in MicroPython. The two neurons are
correspondingly provided with two biases \(b1\). The weights and biases
for the number of neurons in the other layers are coded accordingly:
Three neurons follow in the first hidden layer, followed by two neurons
in the second hidden layer and one neuron in the output layer.

\subsubsection{(Tab. 2) Weights and Biases for 8
Neurons}\label{tab.-2-weights-and-biases-for-8-neurons}

\begin{Shaded}
\begin{Highlighting}[]
\NormalTok{      w1 }\OperatorTok{=}\NormalTok{ [[}\OperatorTok{{-}}\FloatTok{0.75323504}\NormalTok{, }\OperatorTok{{-}}\FloatTok{0.25906014}\NormalTok{],}
\NormalTok{            [}\OperatorTok{{-}}\FloatTok{0.46379513}\NormalTok{, }\OperatorTok{{-}}\FloatTok{0.5019245}\NormalTok{ ],}
\NormalTok{            [ }\FloatTok{2.1273055}\NormalTok{ ,  }\FloatTok{1.7724446}\NormalTok{ ],}
\NormalTok{            [ }\FloatTok{1.1853403}\NormalTok{ ,  }\FloatTok{0.88468695}\NormalTok{]]}
\NormalTok{      b1 }\OperatorTok{=}\NormalTok{ [}\FloatTok{0.53405946}\NormalTok{, }\FloatTok{0.32578036}\NormalTok{]}
\NormalTok{      w2 }\OperatorTok{=}\NormalTok{ [[}\OperatorTok{{-}}\FloatTok{1.6785783}\NormalTok{,  }\FloatTok{2.0158117}\NormalTok{,  }\FloatTok{1.2769054}\NormalTok{],}
\NormalTok{            [}\OperatorTok{{-}}\FloatTok{1.4055765}\NormalTok{,  }\FloatTok{0.6828738}\NormalTok{,  }\FloatTok{1.5902631}\NormalTok{]]}
\NormalTok{      b2 }\OperatorTok{=}\NormalTok{ [ }\FloatTok{1.18362}\NormalTok{  , }\OperatorTok{{-}}\FloatTok{1.1555661}\NormalTok{, }\OperatorTok{{-}}\FloatTok{1.0966455}\NormalTok{]}
\NormalTok{      w3 }\OperatorTok{=}\NormalTok{ [[ }\FloatTok{0.729278}\NormalTok{  , }\OperatorTok{{-}}\FloatTok{1.0240695}\NormalTok{ ],}
\NormalTok{            [}\OperatorTok{{-}}\FloatTok{0.80972326}\NormalTok{,  }\FloatTok{1.4383037}\NormalTok{ ],}
\NormalTok{            [}\OperatorTok{{-}}\FloatTok{0.90892404}\NormalTok{,  }\FloatTok{1.6760625}\NormalTok{ ]]}
\NormalTok{      b3 }\OperatorTok{=}\NormalTok{ [}\FloatTok{0.10695826}\NormalTok{, }\FloatTok{0.01635581}\NormalTok{]}
\NormalTok{      w4 }\OperatorTok{=}\NormalTok{ [[}\OperatorTok{{-}}\FloatTok{0.2019448}\NormalTok{],}
\NormalTok{            [ }\FloatTok{1.5772797}\NormalTok{]]}
\NormalTok{      b4 }\OperatorTok{=}\NormalTok{ [}\OperatorTok{{-}}\FloatTok{1.2177287}\NormalTok{]}
\end{Highlighting}
\end{Shaded}

For pre-training in TensorFlow and Keras the ReLU activation function
was used in the input layer, followed by Tanh in the first hidden layer,
Softmax in the second hidden layer and Sigmoid in the output layer. In
MicroPython (Tab. 3), where users have the option to change the number
of neurons, layers as well as changing the activation functions used,
the corresponding code would be:

\subsubsection{(Tab. 3) Neural Network with 8
Neurons}\label{tab.-3-neural-network-with-8-neurons}

\begin{Shaded}
\begin{Highlighting}[]
\NormalTok{      yout1 }\OperatorTok{=}\NormalTok{ dense(}\DecValTok{2}\NormalTok{, transpose(Xtest), w1, b1, }\StringTok{\textquotesingle{}relu\textquotesingle{}}\NormalTok{)}
\NormalTok{      yout2 }\OperatorTok{=}\NormalTok{ dense(}\DecValTok{3}\NormalTok{, yout1, w2, b2, }\StringTok{\textquotesingle{}tanh\textquotesingle{}}\NormalTok{)}
\NormalTok{      yout3 }\OperatorTok{=}\NormalTok{ dense(}\DecValTok{2}\NormalTok{, yout2, w3, b3, }\StringTok{\textquotesingle{}softmax\textquotesingle{}}\NormalTok{)}
\NormalTok{      ypred }\OperatorTok{=}\NormalTok{ dense(}\DecValTok{1}\NormalTok{, yout3, w4, b4,}\StringTok{\textquotesingle{}sigmoid\textquotesingle{}}\NormalTok{)}
\end{Highlighting}
\end{Shaded}

The Accuracy of this neural network with 8 neurons is 90\%. A variation
of the number of neurons, layers as well as activation functions with
their weights and biases can influence the accuracy accordingly.
Therefore, AI-ANNE: (A) (N)eural (N)et for (E)xploration enables direct
customization of the code in MicroPython via Thonny.

\subsection{Solution with 6 Neurons}\label{solution-with-6-neurons}

The next neural network differs in the number of neurons, the layers and
the activation functions used. This time there are three neurons in the
input layer (Tab. 4). Accordingly, the weights \(w1\) of the neurons are
coded in three columns and four rows, which result from the four
independent variables of the IRIS dataset. This time three biases \(b1\)
need to be added. The code for the hidden layer and the output layer is
written accordingly, so that there is again one neuron in the output
layer for binary classification.

\subsubsection{(Tab. 4) Weights and Biases for 6
Neurons}\label{tab.-4-weights-and-biases-for-6-neurons}

\begin{Shaded}
\begin{Highlighting}[]
\NormalTok{      w1 }\OperatorTok{=}\NormalTok{ [[ }\FloatTok{0.50914556}\NormalTok{, }\OperatorTok{{-}}\FloatTok{0.18116623}\NormalTok{, }\OperatorTok{{-}}\FloatTok{0.04498423}\NormalTok{],}
\NormalTok{            [ }\FloatTok{0.33949652}\NormalTok{, }\OperatorTok{{-}}\FloatTok{0.42303845}\NormalTok{, }\OperatorTok{{-}}\FloatTok{0.37400272}\NormalTok{],}
\NormalTok{            [}\OperatorTok{{-}}\FloatTok{1.4968083}\NormalTok{ ,  }\FloatTok{1.2034143}\NormalTok{ ,  }\FloatTok{0.95544535}\NormalTok{],}
\NormalTok{            [}\OperatorTok{{-}}\FloatTok{1.344156}\NormalTok{  ,  }\FloatTok{0.39220142}\NormalTok{,  }\FloatTok{1.2244085}\NormalTok{ ]]}
\NormalTok{      b1 }\OperatorTok{=}\NormalTok{ [}\FloatTok{0.83684736}\NormalTok{, }\FloatTok{0.5311056}\NormalTok{ , }\FloatTok{0.7652087}\NormalTok{ ]}
\NormalTok{      w2 }\OperatorTok{=}\NormalTok{ [[}\OperatorTok{{-}}\FloatTok{2.1645586}\NormalTok{ ,  }\FloatTok{1.3892978}\NormalTok{ ],}
\NormalTok{            [ }\FloatTok{0.43439832}\NormalTok{, }\OperatorTok{{-}}\FloatTok{1.8758974}\NormalTok{ ],}
\NormalTok{            [ }\FloatTok{0.92036045}\NormalTok{, }\OperatorTok{{-}}\FloatTok{1.5745732}\NormalTok{ ]]}
\NormalTok{      b2 }\OperatorTok{=}\NormalTok{ [}\FloatTok{0.9615521}\NormalTok{, }\FloatTok{0.4445824}\NormalTok{]}
\NormalTok{      w3 }\OperatorTok{=}\NormalTok{ [[ }\FloatTok{1.6905344}\NormalTok{],}
\NormalTok{            [}\OperatorTok{{-}}\FloatTok{2.6346245}\NormalTok{]]}
\NormalTok{      b3 }\OperatorTok{=}\NormalTok{ [}\FloatTok{0.4316521}\NormalTok{]}
\end{Highlighting}
\end{Shaded}

This time, the ReLU activation function was used during pre-training in
TensorFlow and Keras for the input layer, followed by the Sigmoid
activation function in the hidden layer and in the output layer. This
results in the following code in MicroPython (Tab. 5):

\subsubsection{(Tab. 5) Neural Network with 6
Neurons}\label{tab.-5-neural-network-with-6-neurons}

\begin{Shaded}
\begin{Highlighting}[]
\NormalTok{      yout1 }\OperatorTok{=}\NormalTok{ dense(}\DecValTok{3}\NormalTok{, transpose(Xtest), w1, b1, }\StringTok{\textquotesingle{}relu\textquotesingle{}}\NormalTok{)}
\NormalTok{      yout2 }\OperatorTok{=}\NormalTok{ dense(}\DecValTok{2}\NormalTok{, yout1, w2, b2, }\StringTok{\textquotesingle{}sigmoid\textquotesingle{}}\NormalTok{)}
\NormalTok{      ypred }\OperatorTok{=}\NormalTok{ dense(}\DecValTok{1}\NormalTok{, yout2, w3, b3, }\StringTok{\textquotesingle{}sigmoid\textquotesingle{}}\NormalTok{)}
\end{Highlighting}
\end{Shaded}

This results in an Accuracy of 95\%, whereby flexible adjustments are
also possible here. The different Accuracies of the two examples
demonstrate the importance of the number of neurons and layers as well
as the activation functions used. In this case, a simpler neural network
seems to be more appropriate. \newpage

\section{(V) Summary}\label{v-summary}

This working paper explored the integration of neural networks onto
resource-limited microcontrollers and embedded systems like the
Raspberry Pi Pico and Raspberry Pi Pico 2, using a TinyML approach. This
method enabled the deployment of neural networks directly onto
microcontrollers, providing real-time, low-latency, and energy-efficient
inference while ensuring data privacy. For this purpose, AI-ANNE: (A)
(N)eural (N)et for (E)xploration was introduced, a open source framework
that facilitated the transfer of pre-trained models from
high-performance platforms like TensorFlow and Keras to
microcontrollers, using MicroPython as lightweight and transparent
programming languages. The approach demonstrated how key neural network
components --- such as neurons, layers, density, and activation
functions --- could be implemented in MicroPython in order to address
the computational constraints of microcontrollers and embedded systems.
Two neural network examples were presented, either with 8 neurons in 4
layers or with 6 neurons in 3 layers.

Overall, this working paper offered a simple and practical method for
deploying neural networks on energy-efficient devices like
microcontrollers and how they can be used for practical use or as an
educational tool with insights into the underlying technology and
programming techniques of deep learning models. AI-ANNE: (A) (N)eural
(N)et for (E)xploration is an open source project.

\section{About the Author}\label{about-the-author}

\begin{itemize}
\tightlist
\item
  Dennis Klinkhammer completed his doctorate and habilitation at Justus
  Liebig University Giessen (JLU) and focuses his teaching and research
  as a social data scientist on the methodological foundations of
  machine learning and deep learning in the programming languages R and
  Python. Insights into his teaching and research are available at:
  \url{https://www.statistical-thinking.de/}
\end{itemize}

\section{Open Source Code}\label{open-source-code}

\begin{itemize}
\tightlist
\item
  Both the presented and other examples of AI-ANNE: (A) (N)eural (N)et
  for (E)xploration are available on GitHub. This includes Jupyter
  Notebooks for pre-training with TensorFlow and Keras in Python, as
  well as the parameters to be transferred and the corresponding code in
  MicroPython. The link to the repository is:
  \url{https://github.com/statistical-thinking/KI.ENNA/}
\end{itemize}

\section{Sources}\label{sources}

\begin{itemize}
\item
  Cioffi, R.; Travaglioni, M.; Piscitelli, G.; Petrillo, A. and F. de
  Felice (2020): Artificial intelligence and machine learning
  applications in smart production: Progress trends and directions. In:
  Sustainability, 12(2), p.~492.
  \url{https://doi.org/10.3390/su12020492}
\item
  Delnevo, G.; Mirri, S.; Prandi, C.; Manzoni, P. (2023): An evaluation
  methodology to determine the actual limitations of a TinyML-based
  solution. In: Internet of Things, 22, p.~100729.
  \url{https://doi.org/10.1016/j.iot.2023.100729}
\item
  LeCun, Y.; Bengio, Y.; Hinton, G. (2015): Deep Learning. In: Nature,
  521, pp.~436-444. \url{https://doi.org/10.1038/nature14539}
\item
  Ray, P. (2022): A review on TinyML: State-of-the-art and prospects.
  In: Journal of King Saud University - Computer and Information
  Sciences, 34(4), pp.~1595-1623.
  \url{https://doi.org/10.1016/j.jksuci.2021.11.019}
\item
  Sakr, F.; Berta, R.; Doyle, J.; de Gloria, A and F. Bellotti (2021):
  Self-Learning Pipeline for Low-Energy Resource-Constrained Devices.
  In: Energies, 14(20), p 6636. \url{https://doi.org/10.3390/en14206636}
\item
  Wulfert, L.; Kühnel, J.; Krupp, L.; Viga, J.; Wiede, Ch.; Gembaczka,
  P.; Grabmaier, A. (2024): AIfES: A Next-Generation Edge AI Framework.
  In: IEEE Transactions on Pattern Analysis and Machine Intelligence,
  46(6), pp.~4519-4533. \url{https://doi.org/10.1109/TPAMI.2024.3355495}
  \newpage
\end{itemize}

\section{Appendix}\label{appendix}

\subsection{(A) Mathematical Basics in
MicroPython}\label{a-mathematical-basics-in-micropython}

These mathematical basics enable matrix multiplication and other
important operations for the neural network architecture. The codes
(Tab. A) therefore do not need to be adapted in MicroPython!

\subsubsection{(Tab. A) Mathematical
Basics}\label{tab.-a-mathematical-basics}

\begin{Shaded}
\begin{Highlighting}[]
      \CommentTok{\# Mathematical Basics {-} I}
      \KeywordTok{def}\NormalTok{ zero\_dim(x):}
\NormalTok{          z }\OperatorTok{=}\NormalTok{ [}\DecValTok{0} \ControlFlowTok{for}\NormalTok{ i }\KeywordTok{in} \BuiltInTok{range}\NormalTok{(}\BuiltInTok{len}\NormalTok{(x))]}
          \ControlFlowTok{return}\NormalTok{ z}
      
      \CommentTok{\# Mathematical Basics {-} II}
      \KeywordTok{def}\NormalTok{ add\_dim(x, y):}
\NormalTok{          z }\OperatorTok{=}\NormalTok{ [x[i] }\OperatorTok{+}\NormalTok{ y[i] }\ControlFlowTok{for}\NormalTok{ i }\KeywordTok{in} \BuiltInTok{range}\NormalTok{(}\BuiltInTok{len}\NormalTok{(x))]}
          \ControlFlowTok{return}\NormalTok{ z}
      
      \CommentTok{\# Mathematical Basics {-} III}
      \KeywordTok{def}\NormalTok{ zeros(rows, cols):}
\NormalTok{          M }\OperatorTok{=}\NormalTok{ []}
          \ControlFlowTok{while} \BuiltInTok{len}\NormalTok{(M) }\OperatorTok{\textless{}}\NormalTok{ rows:}
\NormalTok{              M.append([])}
              \ControlFlowTok{while} \BuiltInTok{len}\NormalTok{(M[}\OperatorTok{{-}}\DecValTok{1}\NormalTok{]) }\OperatorTok{\textless{}}\NormalTok{ cols:}
\NormalTok{                  M[}\OperatorTok{{-}}\DecValTok{1}\NormalTok{].append(}\FloatTok{0.0}\NormalTok{)}
          \ControlFlowTok{return}\NormalTok{ M}
      
      \CommentTok{\# Mathematical Basics {-} IV}
      \KeywordTok{def}\NormalTok{ transpose(M):}
          \ControlFlowTok{if} \KeywordTok{not} \BuiltInTok{isinstance}\NormalTok{(M[}\DecValTok{0}\NormalTok{], }\BuiltInTok{list}\NormalTok{):}
\NormalTok{              M }\OperatorTok{=}\NormalTok{ [M]}
\NormalTok{          rows }\OperatorTok{=} \BuiltInTok{len}\NormalTok{(M)}
\NormalTok{          cols }\OperatorTok{=} \BuiltInTok{len}\NormalTok{(M[}\DecValTok{0}\NormalTok{])}
\NormalTok{          MT }\OperatorTok{=}\NormalTok{ zeros(cols, rows)}
          \ControlFlowTok{for}\NormalTok{ i }\KeywordTok{in} \BuiltInTok{range}\NormalTok{(rows):}
              \ControlFlowTok{for}\NormalTok{ j }\KeywordTok{in} \BuiltInTok{range}\NormalTok{(cols):}
\NormalTok{                  MT[j][i] }\OperatorTok{=}\NormalTok{ M[i][j]}
          \ControlFlowTok{return}\NormalTok{ MT}
      
      \CommentTok{\# Mathematical Basics {-} V}
      \KeywordTok{def}\NormalTok{ print\_matrix(M, decimals}\OperatorTok{=}\DecValTok{3}\NormalTok{):}
          \ControlFlowTok{for}\NormalTok{ row }\KeywordTok{in}\NormalTok{ M:}
              \BuiltInTok{print}\NormalTok{([}\BuiltInTok{round}\NormalTok{(x, decimals) }\OperatorTok{+} \DecValTok{0} \ControlFlowTok{for}\NormalTok{ x }\KeywordTok{in}\NormalTok{ row])}
      
      \CommentTok{\# Mathematical Basics {-} VI}
      \KeywordTok{def}\NormalTok{ dense(nunit, x, w, b, activation):}
\NormalTok{          res }\OperatorTok{=}\NormalTok{ []}
          \ControlFlowTok{for}\NormalTok{ i }\KeywordTok{in} \BuiltInTok{range}\NormalTok{(nunit):}
\NormalTok{              z }\OperatorTok{=}\NormalTok{ neuron(x, w[i], b[i], activation)}
\NormalTok{              res.append(z)}
          \ControlFlowTok{return}\NormalTok{ res}
\end{Highlighting}
\end{Shaded}

\newpage

\subsection{(B) Activation Funtions in
MicroPython}\label{b-activation-funtions-in-micropython}

The following code (Tab. B) demonstrates how the activation functions
Sigmoid, ReLU, Leaky ReLU, Tanh and Softmax can be programmed in
MicroPython. In TensorFlow and Keras these are already pre-programmed,
in MicroPython the programming has to be done manually. Additional
activation functions can be added accordingly.

\subsubsection{(Tab. B) Activation
Functions}\label{tab.-b-activation-functions}

\begin{Shaded}
\begin{Highlighting}[]
      \CommentTok{\# Sigmoid}
      \KeywordTok{def}\NormalTok{ sigmoid(x):}
\NormalTok{          z }\OperatorTok{=}\NormalTok{ [}\DecValTok{1} \OperatorTok{/}\NormalTok{ (}\DecValTok{1} \OperatorTok{+}\NormalTok{ math.exp(}\OperatorTok{{-}}\NormalTok{x[val])) }\ControlFlowTok{for}\NormalTok{ val }\KeywordTok{in} \BuiltInTok{range}\NormalTok{(}\BuiltInTok{len}\NormalTok{(x))]}
          \ControlFlowTok{return}\NormalTok{ z}
        
      \CommentTok{\# ReLU}
      \KeywordTok{def}\NormalTok{ relu(x):}
\NormalTok{          y }\OperatorTok{=}\NormalTok{ []}
          \ControlFlowTok{for}\NormalTok{ i }\KeywordTok{in} \BuiltInTok{range}\NormalTok{(}\BuiltInTok{len}\NormalTok{(x)):}
              \ControlFlowTok{if}\NormalTok{ x[i] }\OperatorTok{\textgreater{}=} \DecValTok{0}\NormalTok{:}
\NormalTok{                  y.append(x[i])}
              \ControlFlowTok{else}\NormalTok{:}
\NormalTok{                  y.append(}\DecValTok{0}\NormalTok{)}
          \ControlFlowTok{return}\NormalTok{ y}
      
      \CommentTok{\# Leaky ReLU}
      \KeywordTok{def}\NormalTok{ leaky\_relu(x, alpha}\OperatorTok{=}\FloatTok{0.01}\NormalTok{):}
\NormalTok{          p }\OperatorTok{=}\NormalTok{ []}
          \ControlFlowTok{for}\NormalTok{ i }\KeywordTok{in} \BuiltInTok{range}\NormalTok{(}\BuiltInTok{len}\NormalTok{(x)):}
              \ControlFlowTok{if}\NormalTok{ x[i] }\OperatorTok{\textgreater{}=} \DecValTok{0}\NormalTok{:}
\NormalTok{                  p.append(x[i])}
              \ControlFlowTok{else}\NormalTok{:}
\NormalTok{                  p.append(alpha }\OperatorTok{*}\NormalTok{ x[i])}
          \ControlFlowTok{return}\NormalTok{ p}
      
      \CommentTok{\# Tanh}
      \KeywordTok{def}\NormalTok{ tanh(x):}
\NormalTok{          t }\OperatorTok{=}\NormalTok{ [(math.exp(x[val]) }\OperatorTok{{-}}\NormalTok{ math.exp(}\OperatorTok{{-}}\NormalTok{x[val])) }\OperatorTok{/}\NormalTok{ (math.exp(x[val])}
            \OperatorTok{+}\NormalTok{ math.exp(}\OperatorTok{{-}}\NormalTok{x[val])) }\ControlFlowTok{for}\NormalTok{ val }\KeywordTok{in} \BuiltInTok{range}\NormalTok{(}\BuiltInTok{len}\NormalTok{(x))]}
          \ControlFlowTok{return}\NormalTok{ t}
      
      \CommentTok{\# Softmax}
      \KeywordTok{def}\NormalTok{ softmax(x):}
\NormalTok{          max\_x }\OperatorTok{=} \BuiltInTok{max}\NormalTok{(x[val])}
\NormalTok{          exp\_x }\OperatorTok{=}\NormalTok{ [math.exp(val }\OperatorTok{{-}}\NormalTok{ max\_x) }\ControlFlowTok{for}\NormalTok{ val }\KeywordTok{in} \BuiltInTok{range}\NormalTok{(}\BuiltInTok{len}\NormalTok{(x))]}
\NormalTok{          sum\_exp\_x }\OperatorTok{=} \BuiltInTok{sum}\NormalTok{(exp\_x)}
\NormalTok{          s }\OperatorTok{=}\NormalTok{ [j }\OperatorTok{/}\NormalTok{ sum\_exp\_x }\ControlFlowTok{for}\NormalTok{ j }\KeywordTok{in}\NormalTok{ exp\_x]}
          \ControlFlowTok{return}\NormalTok{ s}
\end{Highlighting}
\end{Shaded}

\newpage

\subsection{(C) Neurons in MicroPython}\label{c-neurons-in-micropython}

The predefined activation functions Sigmoid, ReLU, Leaky ReLU, Tanh and
Softmax are already pre-programmed for each neuron in MicroPython (Tab.
C). Activation functions that are added independently must be added
accordingly.

\subsubsection{(Tab. C) Neurons}\label{tab.-c-neurons}

\begin{Shaded}
\begin{Highlighting}[]
      \CommentTok{\# Single Neuron}
      \KeywordTok{def}\NormalTok{ neuron(x, w, b, activation):}
      
\NormalTok{          tmp }\OperatorTok{=}\NormalTok{ zero\_dim(x[}\DecValTok{0}\NormalTok{])}
      
          \ControlFlowTok{for}\NormalTok{ i }\KeywordTok{in} \BuiltInTok{range}\NormalTok{(}\BuiltInTok{len}\NormalTok{(x)):}
\NormalTok{              tmp }\OperatorTok{=}\NormalTok{ add\_dim(tmp, [(}\BuiltInTok{float}\NormalTok{(w[i]) }\OperatorTok{*} \BuiltInTok{float}\NormalTok{(x[i][j]))}
                \ControlFlowTok{for}\NormalTok{ j }\KeywordTok{in} \BuiltInTok{range}\NormalTok{(}\BuiltInTok{len}\NormalTok{(x[}\DecValTok{0}\NormalTok{]))])}
      
          \ControlFlowTok{if}\NormalTok{ activation }\OperatorTok{==} \StringTok{"sigmoid"}\NormalTok{:}
\NormalTok{              yp }\OperatorTok{=}\NormalTok{ sigmoid([tmp[i] }\OperatorTok{+}\NormalTok{ b }\ControlFlowTok{for}\NormalTok{ i }\KeywordTok{in} \BuiltInTok{range}\NormalTok{(}\BuiltInTok{len}\NormalTok{(tmp))])}
          \ControlFlowTok{elif}\NormalTok{ activation }\OperatorTok{==} \StringTok{"relu"}\NormalTok{:}
\NormalTok{              yp }\OperatorTok{=}\NormalTok{ relu([tmp[i] }\OperatorTok{+}\NormalTok{ b }\ControlFlowTok{for}\NormalTok{ i }\KeywordTok{in} \BuiltInTok{range}\NormalTok{(}\BuiltInTok{len}\NormalTok{(tmp))])}
          \ControlFlowTok{elif}\NormalTok{ activation }\OperatorTok{==} \StringTok{"leaky\_relu"}\NormalTok{:}
\NormalTok{              yp }\OperatorTok{=}\NormalTok{ relu([tmp[i] }\OperatorTok{+}\NormalTok{ b }\ControlFlowTok{for}\NormalTok{ i }\KeywordTok{in} \BuiltInTok{range}\NormalTok{(}\BuiltInTok{len}\NormalTok{(tmp))])}
          \ControlFlowTok{elif}\NormalTok{ activation }\OperatorTok{==} \StringTok{"tanh"}\NormalTok{:}
\NormalTok{              yp }\OperatorTok{=}\NormalTok{ tanh([tmp[i] }\OperatorTok{+}\NormalTok{ b }\ControlFlowTok{for}\NormalTok{ i }\KeywordTok{in} \BuiltInTok{range}\NormalTok{(}\BuiltInTok{len}\NormalTok{(tmp))])}
          \ControlFlowTok{elif}\NormalTok{ activation }\OperatorTok{==} \StringTok{"softmax"}\NormalTok{:}
\NormalTok{              yp }\OperatorTok{=}\NormalTok{ tanh([tmp[i] }\OperatorTok{+}\NormalTok{ b }\ControlFlowTok{for}\NormalTok{ i }\KeywordTok{in} \BuiltInTok{range}\NormalTok{(}\BuiltInTok{len}\NormalTok{(tmp))])}
          \ControlFlowTok{else}\NormalTok{:}
              \BuiltInTok{print}\NormalTok{(}\StringTok{"Function unknown!"}\NormalTok{)}
      
          \ControlFlowTok{return}\NormalTok{ yp}
\end{Highlighting}
\end{Shaded}

\subsection{(D) German Free Software
License}\label{d-german-free-software-license}

AI-ANNE: (A) (N)eural (N)et for (E)xploration and KI-ENNA: (E)in
(N)euronales (N)etz zum (A)usprobieren may be used by anyone in
accordance with the terms of the German Free Software License. The
German Free Software License (Deutsche Freie Software Lizenz) is a
license of open-source nature with the same flavors as the GNU GPL but
governed by German law. This makes the license easily acceptable to
German authorities. The D-FSL is available in German and in English.
Both versions are equally binding and are available at:
\url{http://www.d-fsl.org/}

\end{document}